\documentclass[english]{cccconf}
\usepackage[comma,numbers,square,sort&compress]{natbib}
\usepackage{epstopdf}
\usepackage{amssymb}
\usepackage{amscd}
\usepackage{amsfonts}
\usepackage{amsmath}
\usepackage{mathrsfs}
\usepackage{epsfig}
\usepackage{graphicx}
\usepackage{subfigure}
\usepackage{color}
\usepackage{tabularx}
\usepackage[subnum]{cases}
\usepackage{lettrine}
\usepackage[colorlinks=True,linkcolor=blue,anchorcolor=blue,citecolor=blue,CJKbookmarks=true]{hyperref}
\usepackage{bm}
\usepackage{enumerate}
\usepackage{booktabs}
\usepackage{caption}
\usepackage{graphicx}
\usepackage{float} 
\usepackage{subfigure}
\usepackage{multirow}
\usepackage{algorithm}
\usepackage{algorithmicx}
\usepackage{algpseudocode}
\usepackage{amsmath} 
\usepackage{cleveref}
\usepackage{pdfpages}
\usepackage{threeparttable}
\usepackage{multirow}

\begin{document}

\title{An Evidential Real-Time Multi-Mode Fault Diagnosis Approach Based on Broad Learning System}

\author{Chen Li\aref{amss},
        Zeyi Liu\aref{amss},
        Limin Wang\aref{hi},
        Minyue Li\aref{hit},
        Xiao He\aref{amss}}

\affiliation[amss]{Department of Automation, Tsinghua University, Beijing, China
        \email{\{yan12, liuzy21\}@mails.tsinghua.edu.cn, hexiao@mail.tsinghua.edu.cn}}
\affiliation[hi]{School of Mechanical and Electrical Engineering, Guangzhou University, Guangzhou, China
        \email{wanglimin0817@163.com}}
\affiliation[hit]{Wujiang Laboratory, Guiyang, China
        \email{philoinfo@gmail.com}}

\maketitle

\begin{abstract}
Fault diagnosis is a crucial area of research in industry. Industrial processes exhibit diverse operating conditions, where data often have non-Gaussian, multi-mode, and center-drift characteristics. Data-driven approaches are currently the main focus in the field, but continuous fault classification and parameter updates of fault classifiers pose challenges for multiple operating modes and real-time settings. Thus, a pressing issue is to achieve real-time multi-mode fault diagnosis in industrial systems. In this paper, a novel approach to achieve real-time multi-mode fault diagnosis is proposed for industrial applications, which addresses this critical research problem. Our approach uses an extended evidence reasoning (ER) algorithm to fuse information and merge outputs from different base classifiers. These base classifiers based on broad learning system (BLS) are trained to ensure maximum fault diagnosis accuracy. Furthermore, pseudo-label learning is used to update model parameters in real-time. The effectiveness of the proposed approach is demonstrated on the multi-mode Tennessee Eastman process dataset.
\end{abstract}


\footnotetext{This work was supported by the National Natural Science Foundation of China under grant 61733009, National Key Research and Development Program of China under grant 2022YFB25031103, and Huaneng Group Science and Technology Research Project. (\emph{Corresponding author: Xiao He.})}

\section{Introduction}

Fault diagnosis plays a crucial role in ensuring the efficiency, stability, and reliability of industrial processes, making it a focal point in both academic research and industrial applications \cite{FL-2022-OE, WL-2022-MSSP}. However, with the development of integrated, scaled, and complex systems, the challenges posed by fault diagnosis in industrial processes are becoming increasingly demanding. Recent advances in computer and sensor technologies have simplified the data acquisition process and given rise to significant developments in data-driven methods for fault diagnosis \cite{ZL-22-TCST}. Practical industrial processes often involve multiple operating modes, which give rise to non-Gaussian, multi-modal, and center-drifting data features. These characteristics pose a challenge for research into fault diagnosis in industrial production \cite{KP-2015-N}. There are significant differences in process characteristics under different production modes. For multi-mode processes, the data features of each mode are inconsistent. In these cases, it will greatly increase the risk of false positives and false negatives in the diagnosis model. Therefore, studying the technology of fault diagnosis under multi-mode conditions is of great significance.

Some methods have been used to achieve multi-mode fault diagnosis tasks, including time-synchronous averaging \cite{SB-11-MSSP} and order tracking \cite{GH-16-RE}. Li \emph{et al}. proposed an adaptive cost function ridge estimation (CFRE) method for extracting fault-related characteristic curves in a time-frequency representation without the need for rotational speed sensors \cite{YL-2022-TIM}. However, the fault features extracted by these methods are only robust to speed fluctuations or white noise interference, but are sensitive to other changes in the operating environment. Su \emph{et al}. proposed a dilated convolution deep belief network-dynamic multi-layer perceptron (DCDBN-DMLP) for recognizing bearing faults under varying operating conditions, which uses dilated convolution deep belief network, multi-layer domain adaptation, and pseudo label technology to address distribution discrepancies between source and target domains \cite{HS-2022-KS}. Li \emph{et al}. proposed the modified auxiliary classifier GAN (MACGAN) as a novel supervised fault diagnosis model for limited data in rotational machinery \cite{WL-2022-AEI}. Moreover, Hanachi \emph{et al}. proposed a hybrid diagnostic framework combining a data-driven multi-mode fault parameter estimation scheme with a fault propagation model to diagnose hidden incipient faults in gas turbine engine components \cite{HH-2019-MSSP}. However, deep learning methods depend on a large amount of feature data from different operating conditions, which is often difficult to obtain in practical engineering. As such, there is a need to develop a fault diagnosis independent of deep learning that is robust to feature data distribution and operating conditions.

In recent years, various neural network structures have emerged due to the development of neural networks. One such feedforward neural network structure is the broad learning system (BLS), which provides generalization capability for function approximation and does not rely on depth \cite{CL-17-TNNLS}. BLS is a universal approximator for continuous functions on a tight set and is known for its fast learning properties, scaling well to different fields. BLS employs a unique incremental update strategy allowing fast updating and adaptation to new data without requiring retraining, making it highly useful in real-time fault diagnosis tasks \cite{XG-21-TC,ZL-23-TITS}. Unlike traditional neural networks, the incremental update approach of BLS can address issues encountered during the training, such as vanishing and exploding gradients. As such, BLS has extensive applications in fault diagnosis \cite{HZ-19-A, YF-21-TIM}.

This paper proposes a real-time multi-mode fault diagnosis approach that combines the BLS model with an extended evidence reasoning (ER) algorithm. The approach is summarized as follows:
\begin{enumerate}
\item An improved BLS model is proposed, which uses incremental updates with pseudo-labels. It adjusts model parameters dynamically during operation, enabling effective handling of varying environmental conditions in real-time.
\item The extended ER algorithm is adopted to fuse prediction results and allocate reasonable weight and reliability of evidence, ensuring the evidence possesses suitable characteristics in different fault scenarios.
\item The proposed method is validated by multi-mode data experiments using the Tennessee Eastman process.
\end{enumerate}

The rest of the paper is organized as follows: Section \ref{preliminary} provides an introduction to basic concepts. Section \ref{proposed} goes into detail about the proposed approach, which includes a BLS model, evidence reasoning algorithm, and adaptive incremental update procedure based on pseudo-label learning. Section \ref{experiment} describes several comparison experiments conducted based on the MMTEP dataset. The final section, Section \ref{conclusion}, presents concluding remarks.

\section{Preliminaries}\label{preliminary}

\subsection{Broad learning system}

The broad learning system (BLS) is a deep-independent neural network structure derived from the random vector functional-link (RVFL) networks \cite{CL-17-TNNLS}. Its simple structure and exceptional computational speed make it ideal for real-time prediction tasks with limited data features. Given measurement data ${X \in \mathbb{R}^{n \times m}}$ and label ${Y \in \mathbb{R}^{n \times l}}$, the following mapping scheme is utilized:
\begin{equation}
\label{features_BLS}
Z_i=\phi({X}W_{e_i}+\beta_{e_i})\in \mathbb{R}^{n \times N_e},
\end{equation}
where $\phi(\cdot)$ denotes the feature mapping function, and $W_{e_i}$ is the weight matrix of the $i$-th randomly generated feature group. Additionally, $\beta_{e_i}$ is the randomly produced bias vector, and $N_e$ is the number of feature nodes in each feature group. The final feature layer can be represented as $Z^{N_f}=[Z_1,Z_2,\dots,Z_{N_f}]$. The direct correlation exists between the random feature layer $Z^{N_f}$ and the original data $X$. Further mapping operations are necessary to expand the information in $X$:
\begin{equation}
\label{enhancement_BLS}
H_j=\xi (Z^{N_f}W_{h_j}+\beta_{h_j})\in \mathbb{R}^{n \times N_g},
\end{equation}
where $\xi(\cdot)$ denotes the enhanced node mapping function, $W_{h_j}$ is the weight randomly generated for the $j$-th enhancement group, $\beta_{h_j}$ is the bias vector randomly produced for the $j$-th enhancement group, $N_g$ is the number of enhanced nodes in each enhancement group. The enhancement layer can be represented as $H^{N_h}=[H_1,H_2,\cdots,H_{N_h}]$. 

From the calculation process, it can be seen that the enhanced nodes are obtained through the calculation of the feature layer $Z^{N_f}$, which indicates that the enhancement layer $H^{N_h}$ is correlated with the original data, and this further extracts information. Assuming $A^Z=[Z^{N_f}|H^{N_h}]$, the mapping relationship between data $X$ and label $Y$ can be expressed as follows:
\begin{equation}
\label{Relationship_XY}
\begin{aligned}
Y &= {A^Z}W_N^{M}\\
       &= [Z^{N_f}|H^{N_h}]W_N^{M} \\
       &= [Z_1,Z_2,\cdots,Z_{N_f}|H_1,H_2,\cdots,H_{N_h}]W_N^{M}.
\end{aligned}
\end{equation}




\subsection{Evidence reasoning algorithm}

The ER algorithm offers a method for logically combining various sources of information, even when faced with uncertainty and inconsistency \cite{YM-06-EJOR}. It enables evidence to have corresponding characteristics in different decision-making scenarios by reasonably allocating its weight and reliability, which ensures accurate fusion between highly conflicting evidence.

Define the framework of discernment (FOD) as follow:
\begin{equation}
F = \{F_0,F_1,F_2,\cdots,F_N\},
\end{equation}
where $F_i$ represents the $i$-th levels of assessment. All inputs transformed into confidence distribution form can be integrated through the ER fusion framework to obtain the levels of assessment. The form can be expressed as follows:
\begin{equation}
    O(F)=\{(F_n,\beta_n),n=0,1,\cdots,N\},
\end{equation}
where \emph{O} represents a conversion model that transforms FOD \emph{F} into confidence distribution. ${\beta_n}$ can be calculated through the ER algorithm, and its form can be summarized as follows:

Firstly, the relative weights and degrees of belief are combined to convert the degrees of belief into basic probability masses using the following equations:
\begin{equation}
m_{n,i}=w_{i}\beta_{n,i}(x),n=0,1,\cdots,N;i=1,\cdots,L,    
\end{equation}
\begin{equation}
\begin{aligned}
m_{F,i}=&1-\sum_{n=1}^{N}m_{n,i}=-w_{i}\sum_{n=1}^{N}\beta_{n,i}(x),
\end{aligned}   
\end{equation}
\begin{equation}
\overline{{{m}}}_{F,i}=1-w_{i}, 
\end{equation}
\begin{equation}
\widetilde{m}_{F,i}=w_{i}\Big(1-\sum_{n=1}^{N}\beta_{n,i}(x)\Big), 
\end{equation}
where $m_{F,i}=\overline{{{m}}}_{F,i}+\widetilde{m}_{F,i}$ and $\sum_{i=1}^{L}w_i=1$. $0 
\leq \sum_{n=1}^{N}\beta_{n,i}(x) \leq 1$ and $0\leq \beta_{n,i}(x)$. $m_{n,i}$ is the basic probability mass of \emph{x} when the result of assessment is considered to be $F_n$. $w_i$ is the relative weight coefficient of the evidence $i$. $m_{F,i}$ is the probability mass not currently assigned to any single level of assessment and can be split into two parts: $\overline{{{m}}}_{F,i}$ and $\widetilde{m}_{F,i}$. The ER algorithm is used to merge the basic probability masses.
\begin{equation}
\begin{aligned}
m_{n,I(i+1)}=&u_{i+1}[m_{n,I(i)}m_{n,i+1}\\
&+m_{n,I(i)}m_{F,i+1}+m_{F,I(i)}m_{n,i+1}],
\end{aligned}
\end{equation}
\begin{equation}
\begin{aligned}
m_{F,I(i)}=\overline{{{m}}}_{F,I(i)}+\stackrel{\sim}{m}_{F,I(i)},
\end{aligned}
\end{equation}
\begin{equation}
\begin{aligned}
\widetilde{m}_{F,I(i+1)}=&u_{i+1}[\widetilde{m}_{F,I(i)}\widetilde{m}_{F,i+1}\\
&+\widetilde{m}_{F,I(i)}\overline{{m}}_{F,i+1} +\overline{m}_{F,I(i)}\widetilde{m}_{F,i+1}], 
\end{aligned}
\end{equation}
\begin{equation}
\begin{aligned}
\overline{{{m}}}_{F,I(i+1)}=u_{i+1}[\overline{{{m}}}_{F,I(i)}\overline{{{m}}}_{F,i+1}], \\
\end{aligned}
\end{equation}
\begin{equation}
u_{i+1}=\left[\begin{array}{c}{{1-\sum_{n=1}^{N}\sum_{t=1,{{t\neq n}}}^{N}m_{n,I(i)}m_{t,i+1}}}\end{array}\right]^{-1},
\end{equation}
\begin{equation}
\beta_{n}=\frac{m_{n,I(L)}}{1-\overline{{{m}}}_{F,I(L)}},
\end{equation}
where $m_{F,I(1)} = m_{F,1},\widetilde{m}_{F,I(1)}=\widetilde{m}_{F,1}, \overline{m}_{F,I(1)}=\overline{m}_{F,1}$. $I(i+1)$ represents the fusion result of the $i$-th time.

\section{The Proposed Approach}\label{proposed}

In this section, an approach based on BLS and an extended ER algorithm is proposed. The approach is similar to ensemble learning in which information fusion techniques are used to combine the results of multiple classifiers to improve the accuracy of a prediction. Additionally, the model parameters are updated incrementally based on dynamic data streams to increase the adaptability of the model. A multi-mode fault diagnosis approach that fuses the results of multiple classifiers across various operating conditions is proposed for the task of fault diagnosis. 

The modeling process for a given classifier $C_k$ based on BLS can be described as follows:
\begin{equation}\label{feature_node}
z_{i,k}=\phi({X}W_{e_{i,k}}+\beta_{e_{i,k}}).
\end{equation}

Let $z^{N_f}=[z_{1,k},z_{2,k},\cdots,z_{N_f,k}]$. After completing one iteration of the random weight vector mapping, it is advantageous to conduct another iteration to refine the classifier and gain additional knowledge from the input data. Through this iterative process, the classifier can become more accurate and robust to sources of noise or variability in the data.
\begin{equation}
\label{enhancement_node}
h_{j,k}=\xi (z^{N_f}W_{h_{j,k}}+\beta_{h_{j,k}}).
\end{equation}

Denote $h^{N_h}=[h_{1,k},h_{2,k},\cdots,h_{N_h,k}]$. Similar to the process of computing the pseudo-inverse using BLS, it is necessary to flatten and combine the results of two rounds of random mappings $A_{p,k}=[z_{k}^{N_f}|h_{k}^{N_h}]$. The weight parameters of classifier $C_k$ can be obtained as follows:
\begin{equation}\label{Weights_k}
W^M_{N,k} = A_{p,k}^\dag Y.
\end{equation}

The prediction for the data stream \emph{x(t)} at \emph{t} time can be calculated based on Eqs. (\ref{feature_node})-(\ref{Weights_k}) as:
\begin{equation}
    \hat{Y_k}(t) =  A_k(t)W^M_{N,k} = [\hat{y}_{1,k}(t),\hat{y}_{2,k}(t),\cdots,\hat{y}_{N,k}(t)].
\end{equation}

After modeling \emph{K} classifiers, the fusion process of their prediction results at \emph{t} time based on the extended ER algorithm is as follows:
\begin{equation}
m_{n,k}=w_{n,k}\hat{y}_{n,k}(t),n=0,1,\cdots,N;k=1,\cdots,K,    
\end{equation}
\begin{equation}
\begin{aligned}
m_{n,k}^F&=1-\sum_{n=1}^{N}m_{n,k}=1-w_{n,k}\sum_{m=1}^{N}\hat{y}_{m,k}(t), 
\end{aligned}   
\end{equation}
\begin{equation}
\overline{m}^F_{n,k}=1-w_{n,k}, 
\end{equation}
\begin{equation}
\begin{aligned}
\widetilde{m}^F_{n,k}=&w_{n,k}\Big(1-\sum_{n=1}^{N}\hat{y}_{n,k}(t)\Big),
\end{aligned} 
\end{equation}

\begin{equation}
\begin{aligned}
m_{n,I(k+1)}=&u_{k+1}[m_{n,I(k)}m_{n,k+1}\\
&+m_{n,I(k)}m^F_{n,k+1}+m^F_{n,I(k)}m_{n,k+1}],
\end{aligned}
\end{equation}
\begin{equation}
\begin{aligned}
m^F_{n,I(k)}=\overline{m}^F_{n,I(k)}+\stackrel{\sim}{m}^F_{n,I(k)},
\end{aligned}
\end{equation}
\begin{equation}
\begin{aligned}
\widetilde{m}^F_{n,I(k+1)}=&u_{k+1}[\widetilde{m}^F_{n,I(k)}\widetilde{m}^F_{n,k+1}\\
&+\widetilde{m}^F_{n,I(k)}\overline{m}^F_{n,k+1} +\overline{m}^F_{n,I(k)}\widetilde{m}^F_{n,k+1}], 
\end{aligned}
\end{equation}
\begin{equation}
\begin{aligned}
\overline{m}^F_{n,I(k+1)}=u_{k+1}[\overline{m}^F_{n,I(k)}\overline{m}^F_{n,k+1}], \\
\end{aligned}
\end{equation}
\begin{equation}
u_{I(k+1)}=\left[\begin{array}{c}{{1-\sum_{n=1}^{N}\sum_{t=1,{{t\neq n}}}^{N}m_{n,I(k)}m_{t,k+1}}}\end{array}\right]^{-1},
\end{equation}
\begin{equation}
\hat{y}^e_{n}(t)=\frac{m_{n,I(K)}}{1-\overline{m}^F_{n,I(K)}}.
\end{equation}

The results after fusion using the extended ER algorithm can be obtained as $\hat{Y}^e(t)=[\hat{y}^e_{1}(t),\hat{y}^e_{2}(t),\cdots,\hat{y}^e_{N}(t)]$. Note that any classifier $C_k$ has a corresponding relative weight $w_{n,k}$ in each system state \emph{k}. $m^F_{n,I(1)} = m^F_1,\widetilde{m}^F_{n,I(1)}=\widetilde{m}^F_{n,1}, \overline{m}^F_{n,I(1)}=\overline{m}^F_{n,1},n = 1,2,\cdots,N$. $I(k+1)$ is the fusion result of the $k$-th time.
\begin{figure*}[!ht]
\centering
\subfigure[Mode 1.]{
\begin{minipage}[t]{0.33\linewidth}
\centering
\includegraphics[width=1\linewidth]{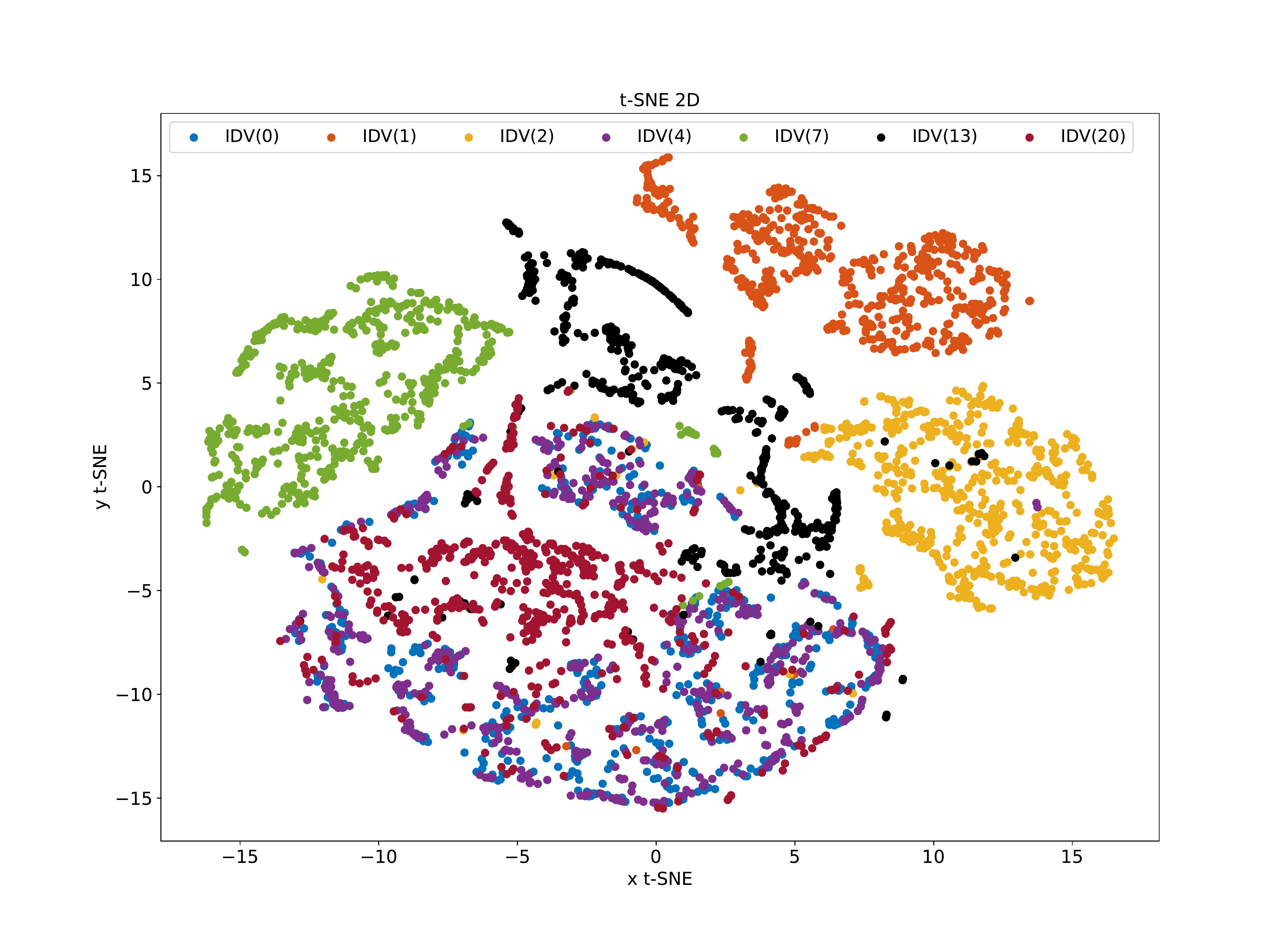}
\label{M1_tsne}
\end{minipage}%
}%
\subfigure[Mode 2.]{
\begin{minipage}[t]{0.33\linewidth}
\centering
\includegraphics[width=1\linewidth]{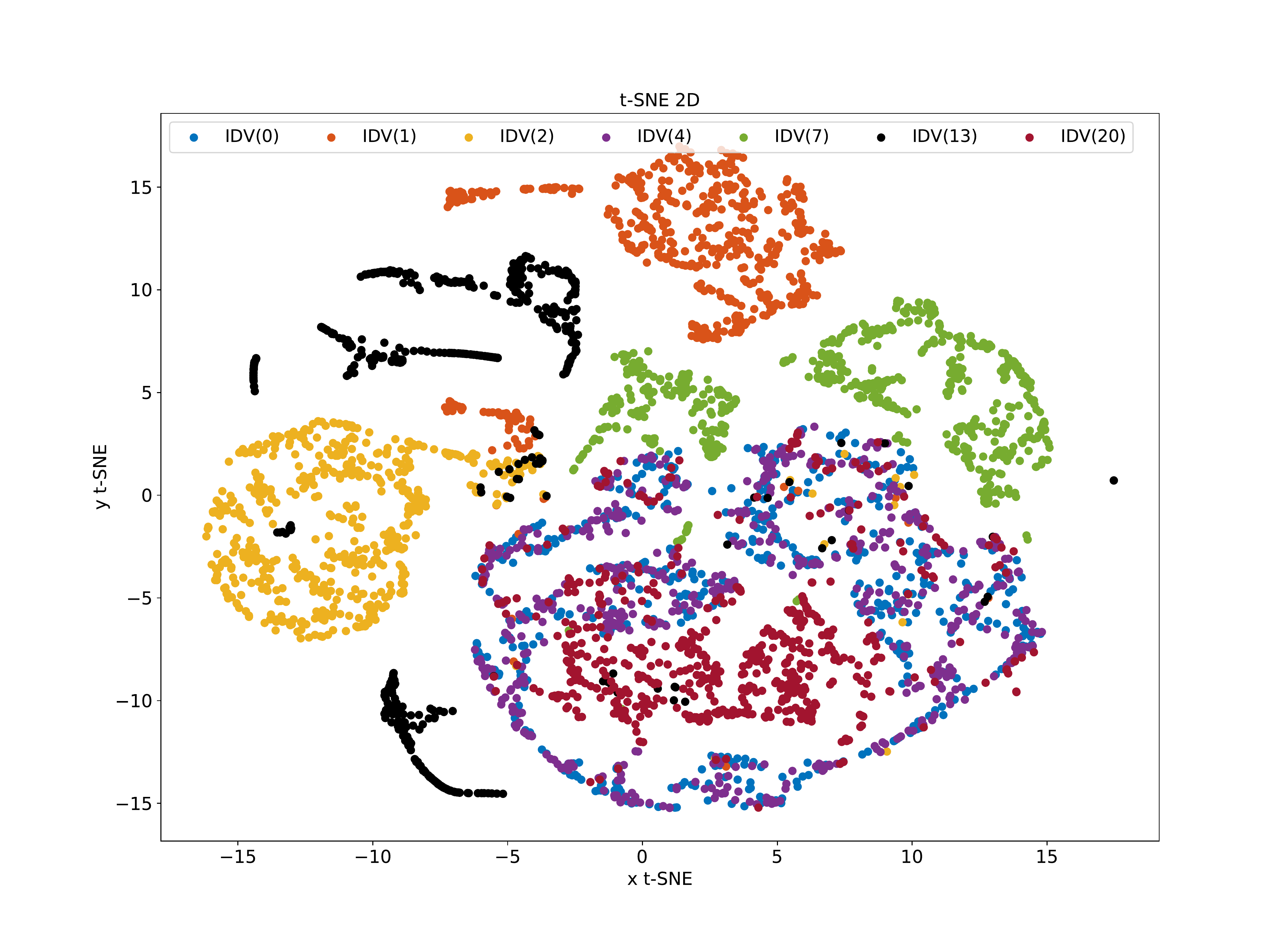}
\label{M2_tsne}
\end{minipage}%
}%
\subfigure[Mode 3.]{
\begin{minipage}[t]{0.33\linewidth}
\centering
\includegraphics[width=1\linewidth]{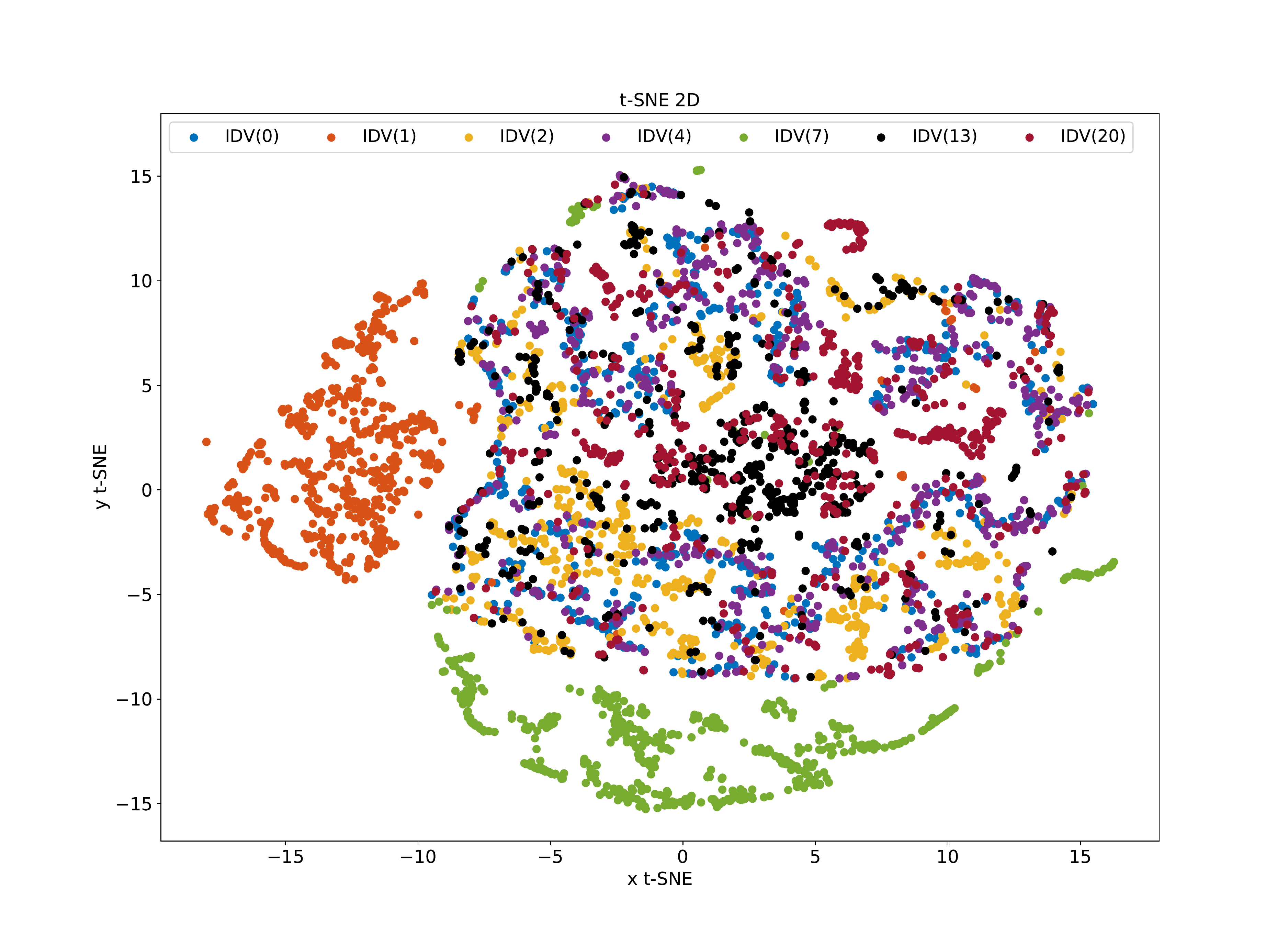}
\label{M3_tsne}
\end{minipage}
}%
\\
\subfigure[Mode 4.]{
\begin{minipage}[t]{0.33\linewidth}
\centering
\includegraphics[width=1\linewidth]{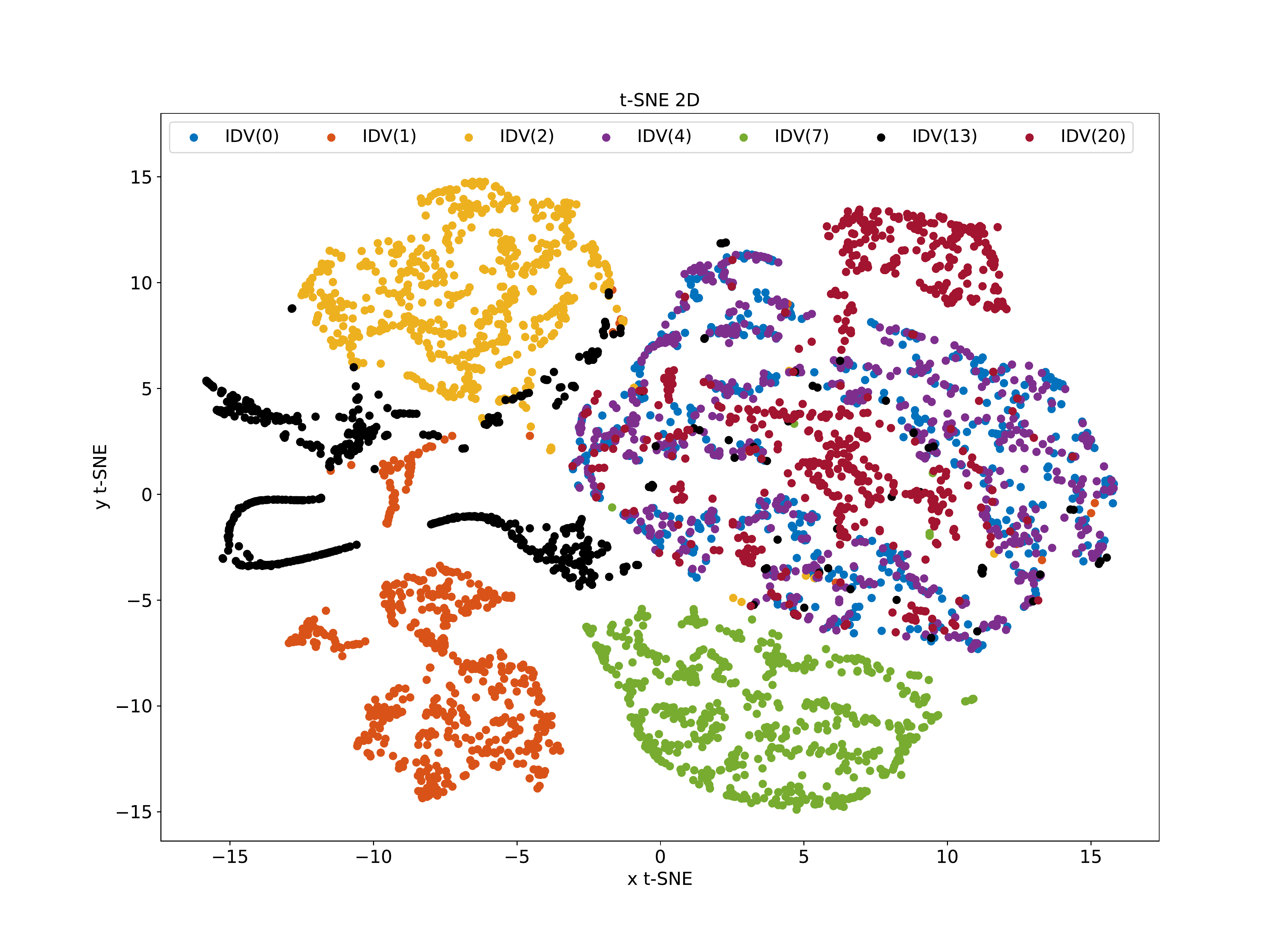}
\label{M4_tsne}
\end{minipage}%
}%
\subfigure[Mode 1 and Mode 2.]{
\begin{minipage}[t]{0.33\linewidth}
\centering
\includegraphics[width=1\linewidth]{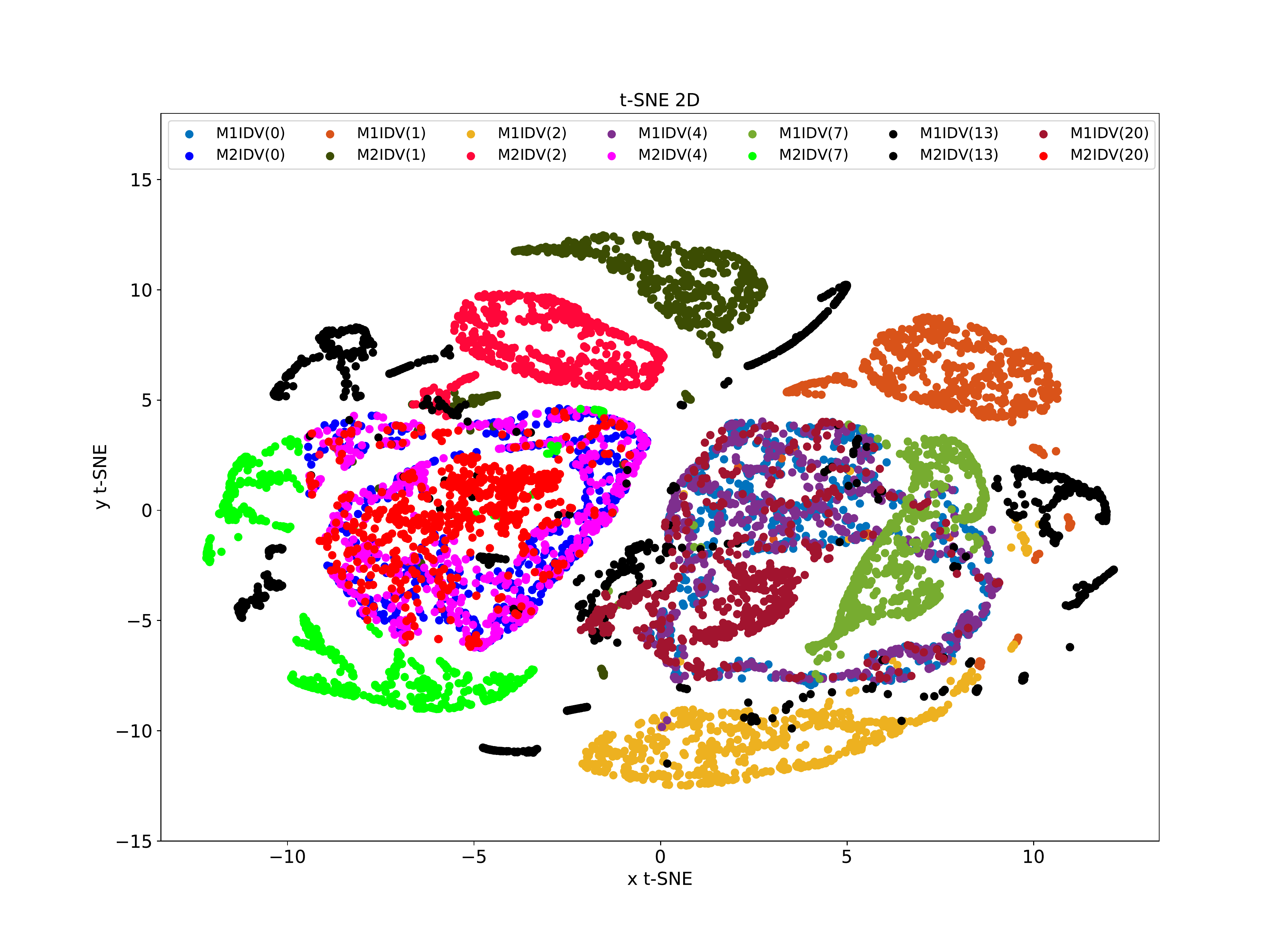}
\label{M1M2_tsne}
\end{minipage}
}%
\subfigure[Mode 3 and Mode 4.]{
\begin{minipage}[t]{0.33\linewidth}
\centering
\includegraphics[width=1\linewidth]{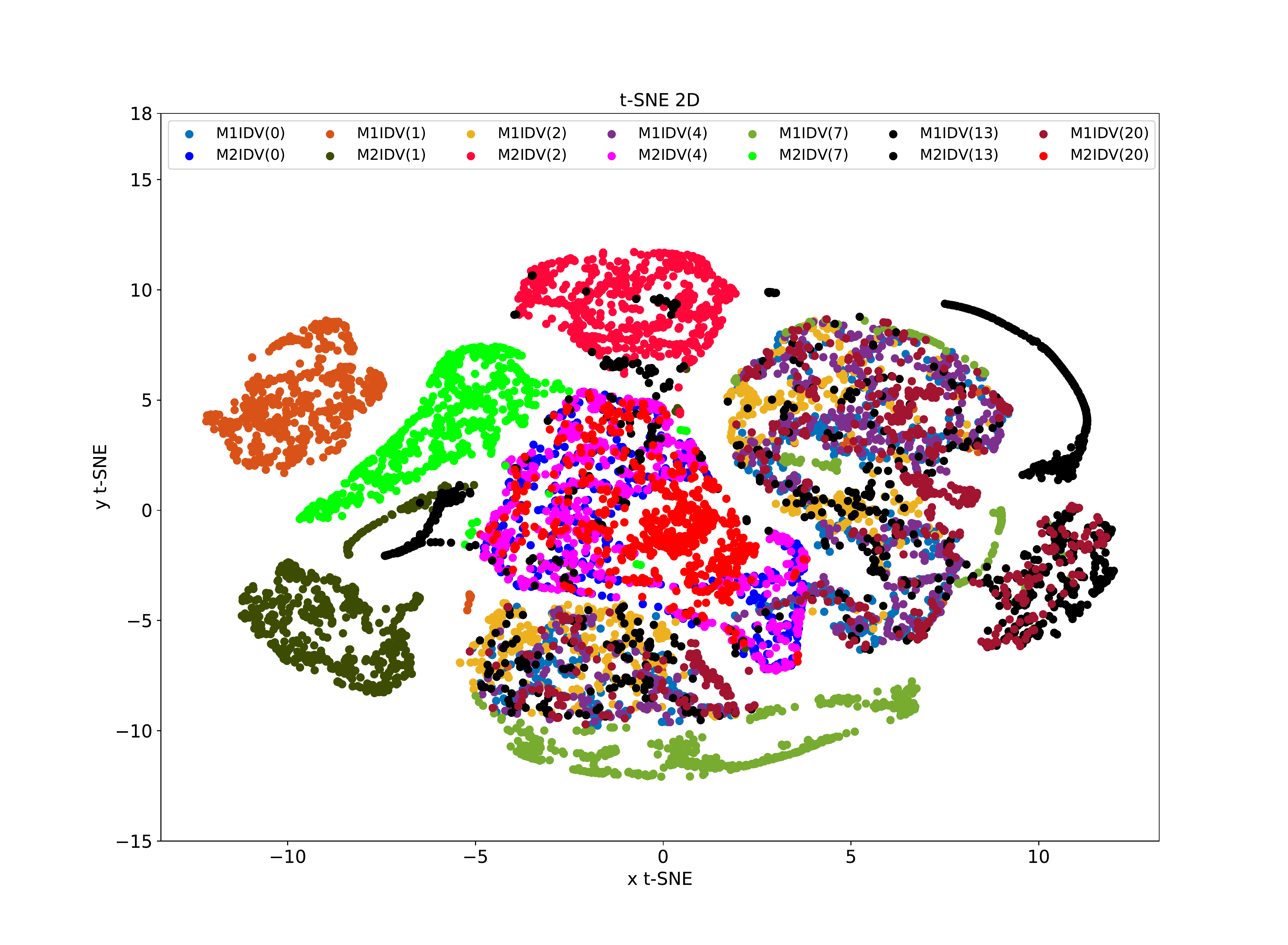}
\label{M5_tsne}
\end{minipage}%
}%
\caption{The t-SNE visualization results of six faults in different modes.}
\label{visualization_results_faults}
\end{figure*}


The actual system operation process is non-stationary, which can cause changes in the underlying components of the data \cite{ZL-22-TNNLS,ZL-21-TIM}. The pseudo-labels (PL) technique is a semi-supervised learning tool that utilizes both labeled and unlabeled data in a supervised manner \cite{DH-12-ICML}. With respect to the unlabeled data, PL selects the class with the highest predicted possibility during each weight update, treating this class as if it were the true label. In most cases, this method is compatible with almost any neural network model or training methodology. As for the new incoming data $x(t)$ at time \emph{t}, it can obtained as follows:
\begin{equation}
z_{i,k}(t)=\phi({x(t)}W_{e_{i,k}}+\beta_{e_{i,k}}),
\end{equation}
\begin{equation}
h_{j,k}(t)=\xi (z^{N_f}(t)W_{h_{j,k}}+\beta_{h_{j,k}}),
\end{equation}
\begin{equation}\label{H_newdata}
{A_k(t)} = [z_{k}^{N_f}(t)|h_{k}^{N_h}(t)],
\end{equation}
where $z^{N_f}_k(t)=[z_{1,k}(t),z_{2,k}(t),\cdots,z_{N_f,k}(t)]$ and $h^{N_h}_k(t)=[h_{1,k}(t),h_{2,k}(t),\cdots,h_{N_h,k}(t)]$. Let $A_{k,l} = A_{p,k}$. Therefore, the extended matrix can be obtained as:
\begin{equation}
    A_{k,l+1}=\left[\begin{array}{l}
         A_{k,l}\\
         A_k(t)
    \end{array} 
    \right].
\end{equation}

Next, the related pseudo-inverse update algorithm can be derived:
\begin{equation}\label{pseudoinverse_update_bewdata}
    A_{k,l+1}^\dag=[A_{k,l}^\dag-B_kD_k^T|B_k],
\end{equation}
where $D_k^T=A_k(t)A_{k,l}^\dag$,
\begin{eqnarray}\label{Transformtopesudolabel}
{B_k^T}  = 
  \begin{cases}
    C_k^\dag, & {\textit{if}} \ {C_k \ne 0},\\\hfill
    {{{\left( {1 + {D_k^T} {D_k}} \right)}^{ - 1}}{A}_{k,l}^\dag{D_k} }, & \textit{if}~{C_k = 0},
  \end{cases}
\end{eqnarray}

\begin{equation}
    C_k=A_k(t)-D_k^TA_{k,l}.
\end{equation}
\begin{figure}[!ht]
\centering
\includegraphics[width=1\linewidth]{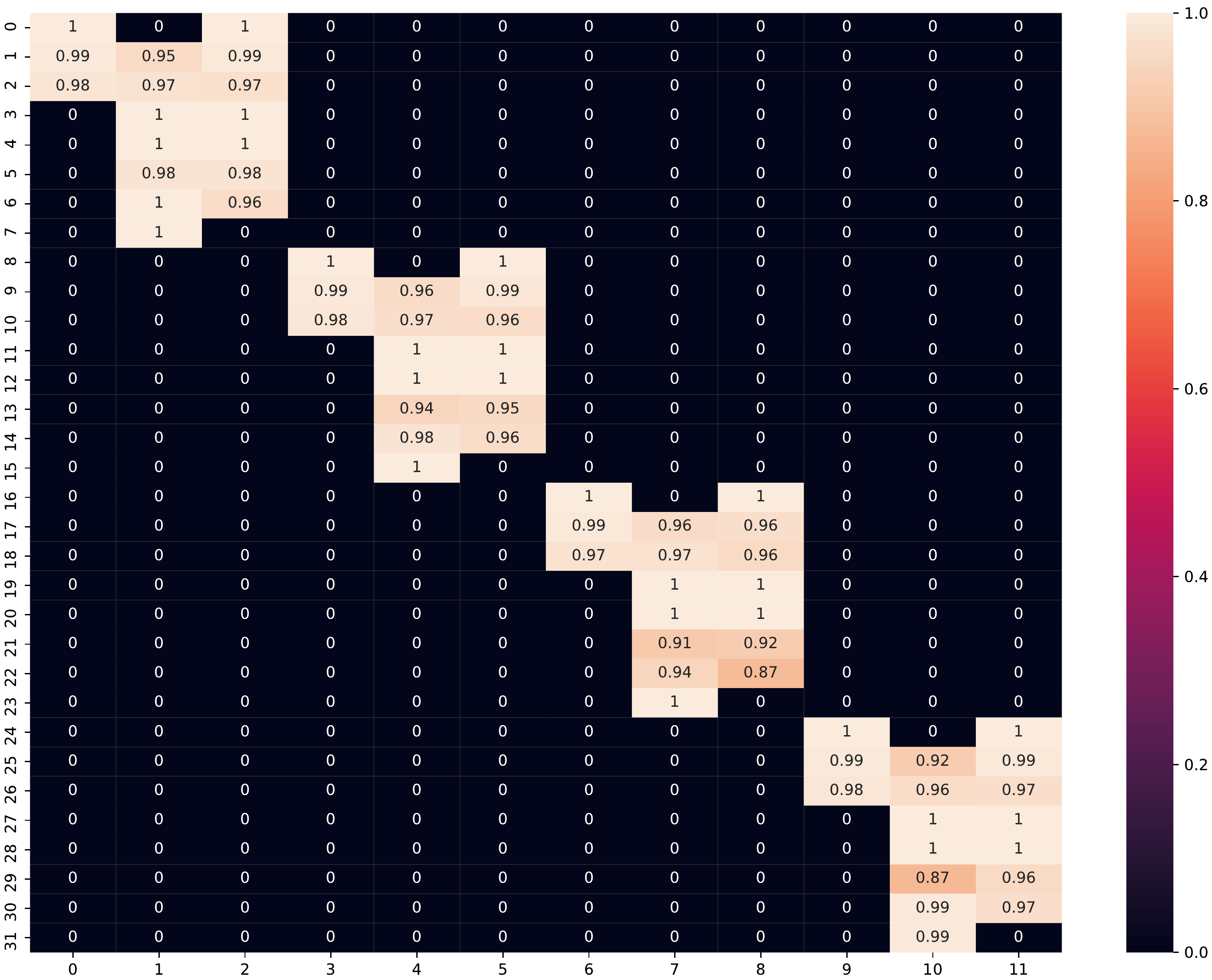}
\caption{The weights of base classifiers in different states of each mode.}
\label{weights_mode}
\end{figure}

The weight after the input data $x(t)$ is updated as:
\begin{equation}\label{weights_updata_newdata}
    W_{k,l+1}=W_{k,l}+(Y(t)-A_k(t)W_{k,l})B_k,
\end{equation}
where $Y(t)$ is the true label of the new data $x(t)$. However, it is actually difficult to obtain true labels in real-time processes. The pseudo-labels generated from the model's predictive results are used to update the weights:
\begin{equation}
    W_{k,l+1}=W_{k,l}+(\widetilde Y(t)-A_k(t)W_{k,l})B_k,
\end{equation}
where $\widetilde Y(t) = [\widetilde y_{1}(t),\widetilde y_{2}(t),\cdots,\widetilde y_{N}(t)]$ is the pseudo-label generated from the model's predictive results. The way of pseudo-label learning can be summarized as:
\begin{eqnarray}
{\widetilde y_{i}(t)} = 
  \begin{cases}

     1, & {\textit{if}} \ \ {i} = \mathop {\arg\max }\limits_i~ \hat{Y}^e(t),\\\hfill

    0, & \textit{otherwise}.

  \end{cases}
\end{eqnarray}

For unlabelled data, pseudo-labels only pick the class with the highest predicted probability at each weight update and utilize them as if they were true labels.

\begin{figure*}[!ht]
\centering
\subfigure[Mode 1.]{
\begin{minipage}[t]{0.4\linewidth}
\centering
\includegraphics[width=1\linewidth]{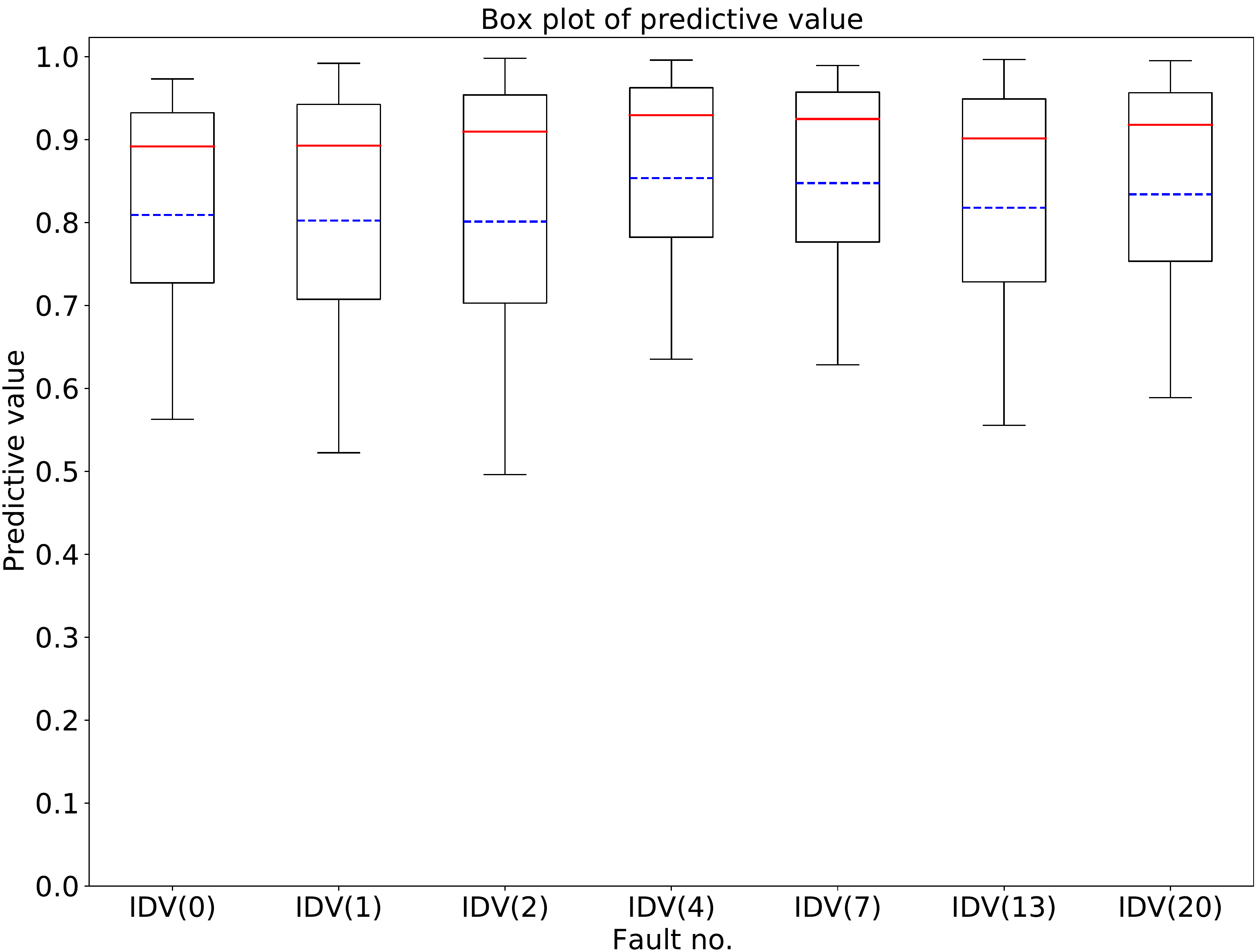}
\label{M1_box}
\end{minipage}%
}%
\subfigure[Mode 2.]{
\begin{minipage}[t]{0.4\linewidth}
\centering
\includegraphics[width=1\linewidth]{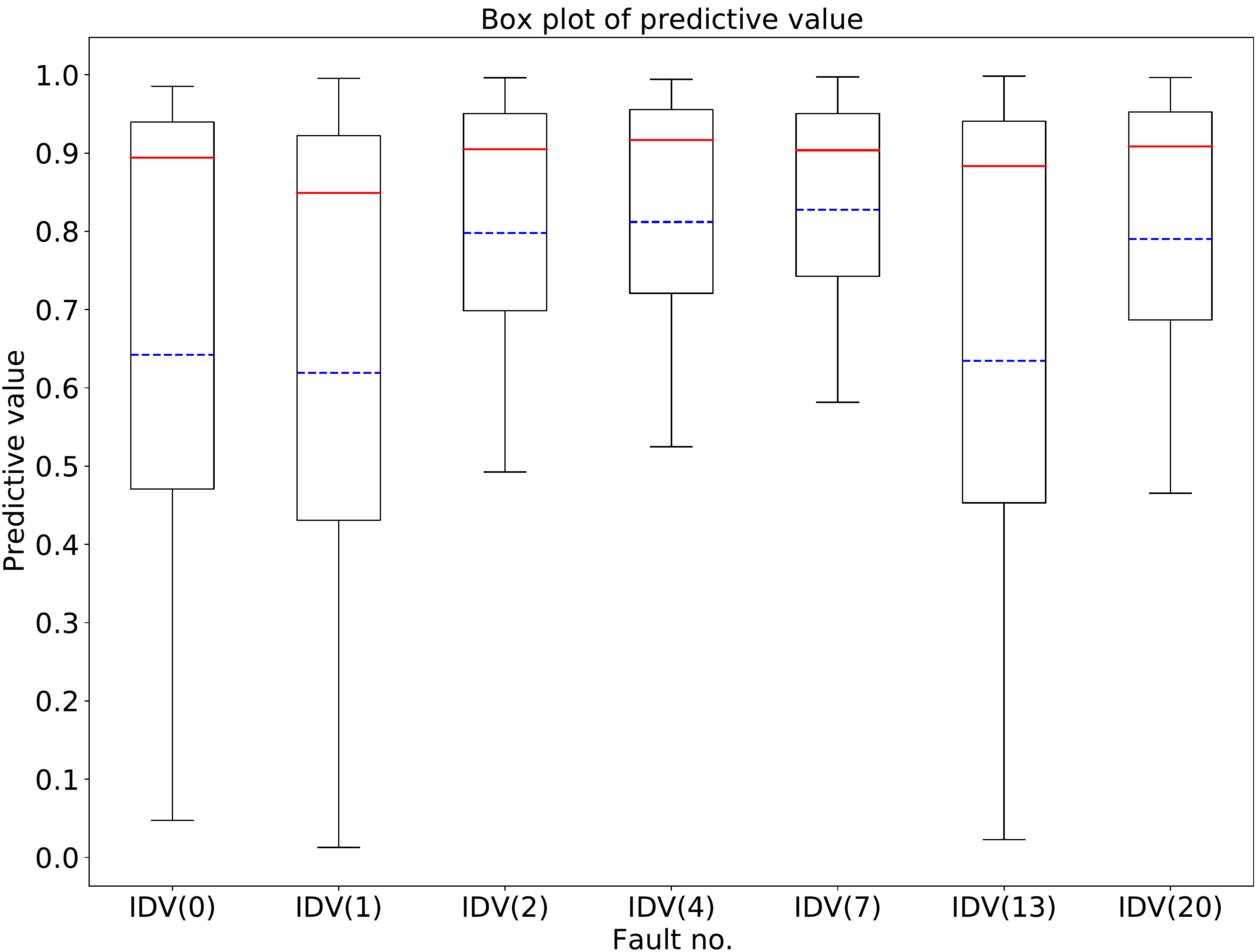}
\label{M2_box}
\end{minipage}%
}%
\\
\subfigure[Mode 3.]{
\begin{minipage}[t]{0.4\linewidth}
\centering
\includegraphics[width=1\linewidth]{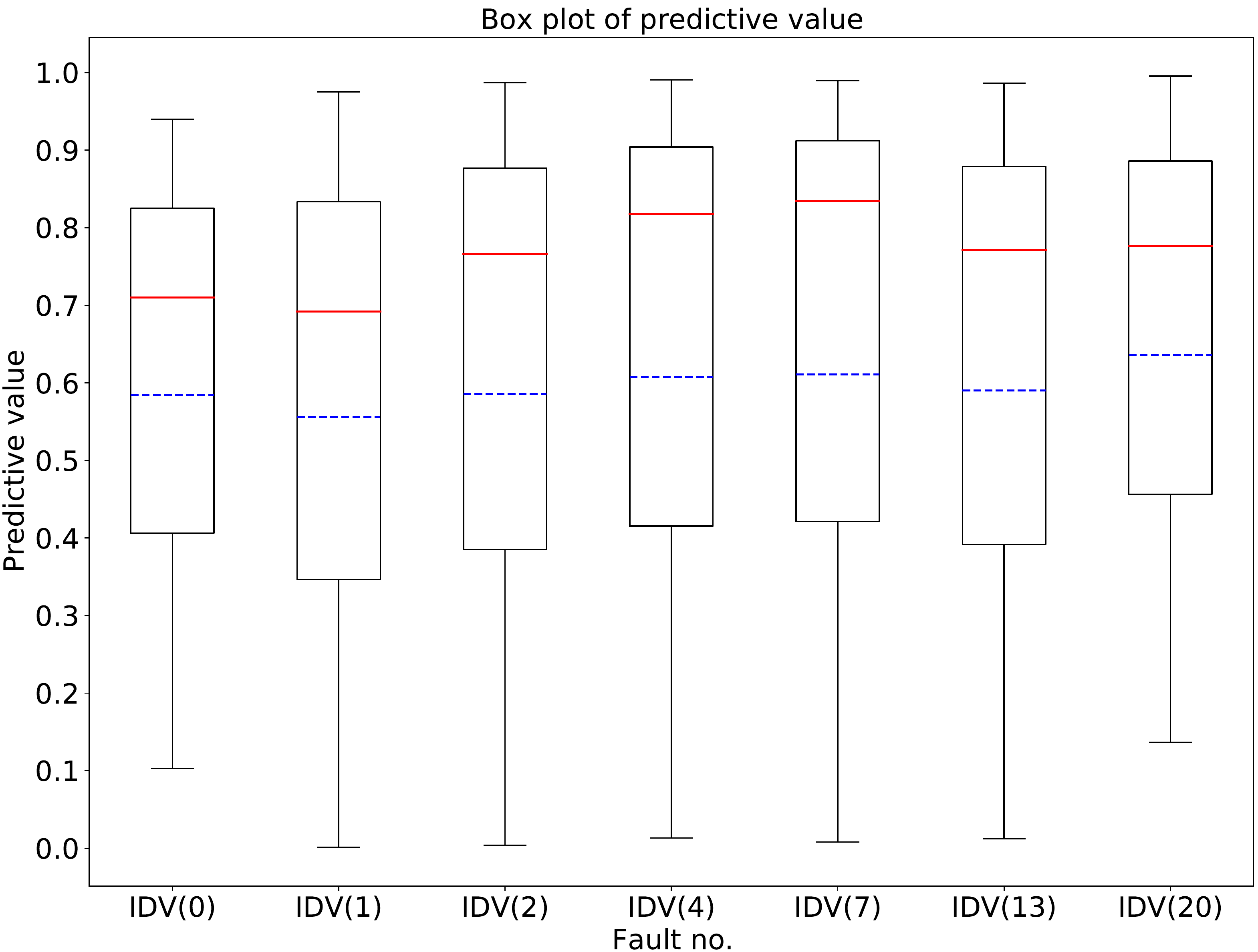}
\label{M3_box}
\end{minipage}
}%
\subfigure[Mode 4.]{
\begin{minipage}[t]{0.4\linewidth}
\centering
\includegraphics[width=1\linewidth]{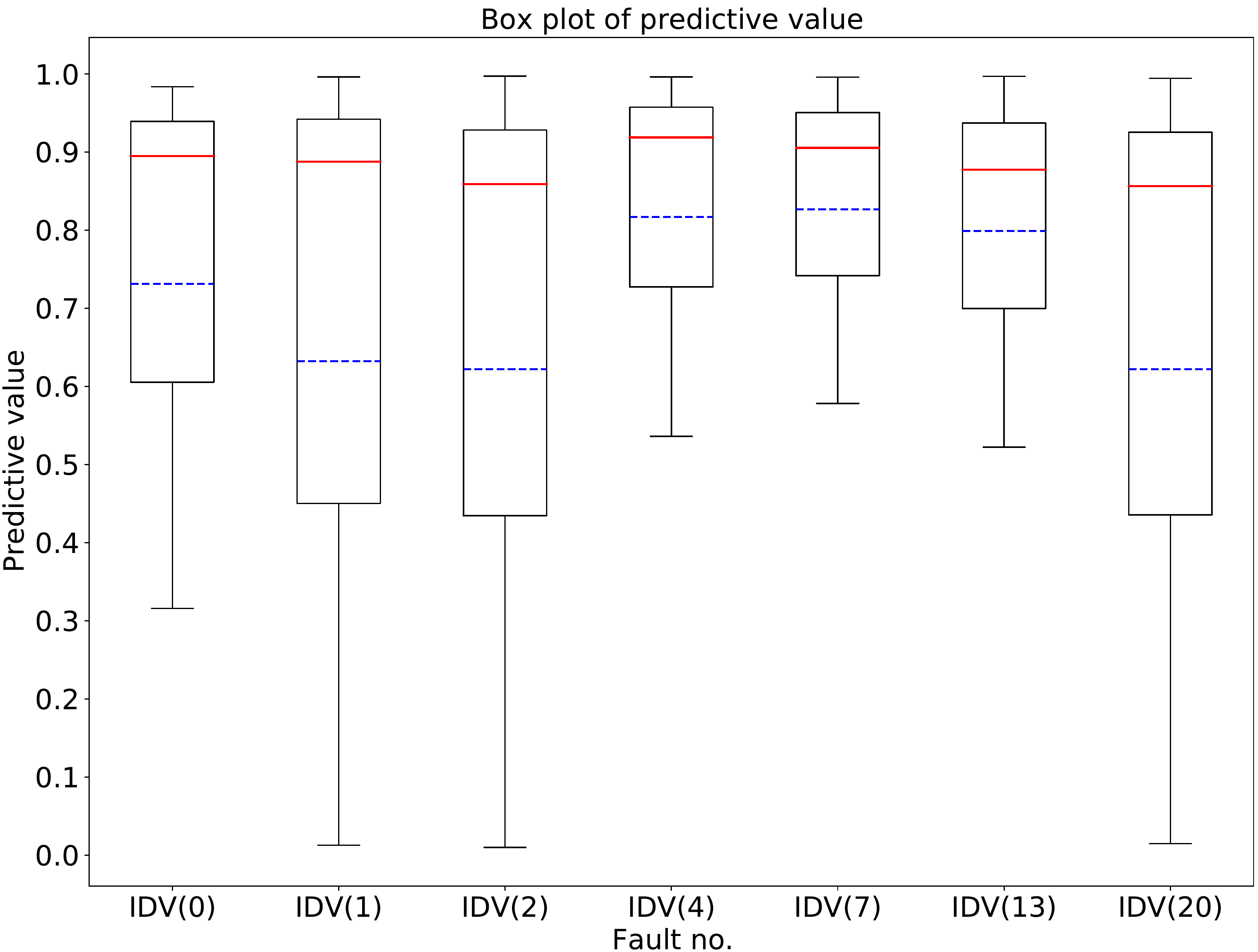}
\label{M4_box}
\end{minipage}%
}%
\caption{The results of the prediction in different modes.}
\label{Boxplot_modes}
\end{figure*}
\begin{table*}
\centering
\caption{Accuracy of different diagnostic schemes (\%)}
\setlength{\tabcolsep}{0.17cm}{
\begin{tabular}{c c c c c c c c c c c c c}
 \specialrule{0.1em}{1pt}{1pt}
 \specialrule{0.1em}{1pt}{3pt}
\multirow{2}{*}{\it  Fault}  &\multicolumn{4}{c}{\it Our Scheme} &\multicolumn{4}{c}{\it Scheme 1} &\multicolumn{4}{c}{\it Scheme 2}\\
\cmidrule(r){2-5}\cmidrule(r){6-9}\cmidrule(r){10-13}
&\it Mode 1      &\it Mode 2     &\it Mode 3    &\it Mode 4      &\it Mode 1      &\it Mode 2     &\it Mode 3    &\it Mode 4     &\it Mode 1      &\it Mode 2     &\it Mode 3    &\it Mode 4    \\
 \specialrule{0.1em}{1pt}{1pt}
IDV(0)  &100    &99.9   &95.2   &99.9   &100    &100   &100   &100   &100    &100   &100   &100\\
IDV(1)  &100    &99.8   &86.7   &99.9   &100    &100   &100   &100    &100    &100   &100   &100\\
IDV(2)  &100    &100    &96.6   &99.7   &100    &100   &100   &100   &100    &100   &100   &100\\
IDV(4)  &100    &100    &99.7   &100   &100    &85.7  &0     &98.4    &100    &100   &100   &100\\
IDV(7)  &100    &100    &99.6   &100   &100    &100   &100   &100    &100    &100   &100   &100\\
IDV(13) &100    &99.7   &90.6   &100    &92.9   &82.1  &0     &100   &100    &100   &35.4   &100\\
IDV(20) &100    &100    &94.8   &99.7    &84.3   &69.1  &38.3  &64.6    &98     &91.5  &39.3   &85.4\\
 \specialrule{0.1em}{1pt}{3pt}
 \specialrule{0.1em}{1pt}{1pt}
\end{tabular}}
\label{Acarracy_scheme}
\end{table*}

\section{Experiments}\label{experiment}

\subsection{Datasets}

This study uses the multi-mode Tennessee Eastman process (MMTEP) dataset to perform experiments. The MMTEP dataset is distinct from traditional datasets as it contains multiple modalities, making it more complex and diverse. Its data streams from various sensors, such as pressure, temperature, and flow sensors, provide different modalities (e.g., bias, lag, etc.). It is based on the Tennessee Eastman process (TEP) dataset and provides an excellent resource for researching multi-mode data analysis and pattern recognition. These sensors collect data from different locations, providing various perspectives on process monitoring and control. The MMTEP dataset is applicable for various research directions, including multi-mode data analysis, multi-mode pattern recognition, and multi-mode fusion\footnote[1]{The dataset is released at \url{https://github.com/THUFDD/Multi-mode-Fault-Diagnosis-Datasets-with-TE-process}}.

Furthermore, the MMTEP dataset provides an experimental dataset for applications such as process monitoring, fault diagnosis, and state estimation in the engineering field. These practical applications contribute to solving real-world engineering problems. There are 41 measured variables, i.e., XMEAS (1–41), and 12 manipulated variables, i.e., XMV (1–12) in MMTEP. And there are 28 faults IDV(1)-(28) and a normal state IDV(0) in dataset. Six faults in four modes, including IDV(1), IDV(2), IDV(4), IDV(7), IDV(13), and IDV(20), are selected as the experimental dataset in this paper. The visualization results of the experimental dataset are shown in Fig. \ref{visualization_results_faults}.

\subsection{Analysis of experimental results}

To address the various types of faults encountered in the experiment, a specific approach is utilized in the experiment. For each mode, we randomly selected six types of fault data from the full set of faults to construct the training set. The remaining data were allocated to the test set.

To further enhance the classification accuracy, three base classifiers based on BLS were trained to achieve classification. The base classifiers were used to accurately differentiate each state in the current mode from other states (including those from other modes), allowing for a more comprehensive evaluation of the model's ability to classify. The extended ER algorithm is used to combine the outputs of the multiple base classifiers, resulting in improved classification accuracy. In addition, the adaptive incremental learning design was used to dynamically adjust the parameters of base classifiers with continuous updates to facilitate the response to new data streams. Fig. \ref{weights_mode} displays the weight of each base classifier for each fault in each mode in the experiment. The results of the prediction are shown in Fig. \ref{Boxplot_modes}.

To verify the effectiveness of the proposed approach, two BLS-based model training schemes were compared:
\begin{itemize}
    \item \emph{Scheme 1}: Train a single classifier using a training dataset containing all modes' data.
    \item \emph{Scheme 2}: Train a classifier for each mode using the data from each mode separately.
\end{itemize}

The classification accuracy of the proposed approach can be found in Table \ref{Acarracy_scheme} which displays that the approach has excellent diagnostic performance in all modalities. It suggests that the combination of wide learning and evidence inference rules can effectively improve the accuracy and robustness of fault diagnosis. The approach can reduce mutual interference between modalities, thus better ensuring the accuracy and reliability of fault diagnosis.

\section{Conclusion}\label{conclusion}

In this paper, an improved approach for the real-time fault diagnosis task in multi-mode processes has been proposed. The approach has incorporated BLS as base classifiers and the extended ER algorithm for result fusion. Additionally, a PL learning method has been used to incrementally update the base classifier parameters. BLS could fully explore the data space and improve feature representation. The extended ER algorithm could effectively combine the outputs of multiple base classifiers and enhance robustness and accuracy. The proposed approach has achieved better fault diagnosis performance in situations where there were multiple operating modes. It has fused the information from several basic classifiers and combined their respective advantages to improve the accuracy of fault diagnosis. The effectiveness of the proposed approach has been verified in the multi-mode Tennessee Eastman process dataset.

\end{document}